\documentclass[10pt,twocolumn]{article}

\usepackage[T1]{fontenc}
\usepackage[utf8]{inputenc}
\usepackage[english]{babel}

\usepackage[protrusion=true,expansion=true]{microtype}
\usepackage{lmodern}

\usepackage{amsmath,amsfonts,amsthm,amssymb}

\usepackage{graphicx}
\usepackage{epstopdf}
\usepackage{subfig}
\usepackage{booktabs}
\usepackage{tabularx}
\usepackage{multirow}
\usepackage{float}
\usepackage{afterpage}

\usepackage[svgnames]{xcolor}
\usepackage[font=small,labelfont=bf,textfont=it]{caption}

\usepackage[numbers,sort&compress]{natbib}

\usepackage{authblk}
\usepackage{newfloat}
\usepackage{comment}
\usepackage{svg}
\usepackage{titlesec}

\DeclareFloatingEnvironment[name={Supplementary Figure}]{suppfigure}

\usepackage[hyperfootnotes=false,hidelinks]{hyperref}

\titlespacing*{\section}{0pt}{0.8ex}{0.3ex}
\titlespacing*{\subsection}{0pt}{0.6ex}{0.2ex}
\titlespacing*{\subsubsection}{0pt}{0.4ex}{0.15ex}

\title{\Large\bfseries
Multimodal AI in Healthcare: Review of Vision-Language Foundation Models for Real-World Medical Applications
}

\author[1,2]{Taha Razzaq\thanks{Corresponding author: 21030019@lums.edu.pk}}
\author[1]{Murtaza Taj}
\author[2]{Asim Iqbal}

\affil[1]{Lahore University of Management Sciences}
\affil[2]{Tibbling Technologies}

\date{}

\begin{document}

\maketitle

\textbf{The emergence of foundation models has marked a transformative shift in AI, enabling robust generalization across diverse downstream tasks through putative zero-shot learning. Large Language Models and Vision-Language Models have demonstrated strong capabilities in tasks such as image interpretation, report generation, and question answering by effectively learning from multimodal data—images paired with associated text—often with minimal supervision. In the healthcare domain, this ability to align visual and textual information reduces the reliance on extensive manual annotations, as models can leverage existing clinical reports and imaging data to learn meaningful representations. This integration holds promise for improving diagnostic support, treatment planning, and overall patient care, even in data-constrained settings. In this review, we provide a definitive taxonomy of the medical VLM landscape, tracing the evolution from early Contrastive Alignment and Generative MLLMs to the cutting-edge frontiers of Dense Pixel-Grounding, Sparse Mixture-of-Experts (MoE), and Reasoning-Incentivized (RL) architectures. We critically examine the ``medical bottleneck"—identifying the persistent challenges of data scarcity, the ``hallucination" risks in generative diagnostics, the computational strain of 3D volumetric processing, and the lack of standardized, clinically-grounded evaluation metrics.
}

\section{Introduction}

The landscape of artificial intelligence has undergone a fundamental shift with the emergence of Foundation Models (FMs), large-scale architectures, typically based on the Transformer paradigm, that are pre-trained on vast, heterogeneous datasets. Unlike traditional deep learning (DL) models that rely on task-specific supervision, FMs leverage self-supervised learning to develop robust, transferable representations. This allows them to excel in downstream tasks through zero-shot or few-shot learning, effectively navigating the "long tail" of rare conditions where labeled data is historically sparse (\cite{survey, fm_survey}). The scalability of these models enables the capture of emergent properties—capabilities that arise only at high computational and data scales—thereby reducing the prohibitive costs of manual annotation that have long bottlenecked AI deployment in specialized fields.
\begin{figure*}[!ht]
    \centering
    \includegraphics[scale=0.5]{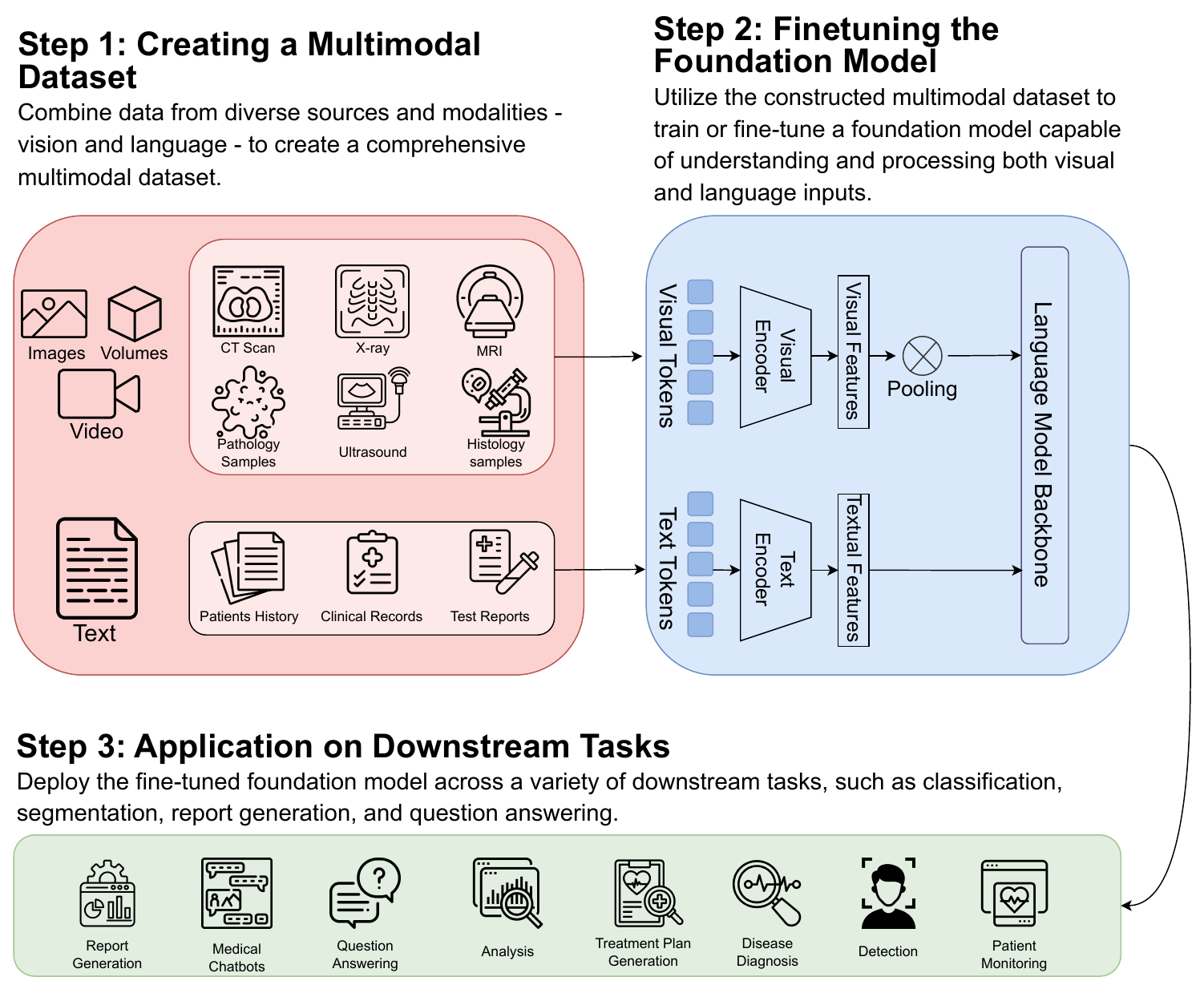}
    \caption{Overview of the foundation model pipeline in the medical domain. \textbf{Step 1} involves building a multimodal dataset by integrating visual and language data. In \textbf{Step 2}, this dataset is used to fine-tune a foundation model capable of processing both modalities. \textbf{Step 3} illustrates the various medical downstream tasks, including feature classification, segmentation, report generation, and question answering, on which the foundation model can be applied.}
    \label{fig:fig1}
\end{figure*}

This surge in FMs was catalyzed by breakthroughs in Large Language Models (LLMs), such as GPT-4 (\cite{floridi2020gpt}) and LLaMA-3 (\cite{llama}). These models demonstrated that massive-scale pre-training on natural language could yield sophisticated reasoning and context-awareness. However, human intelligence and professional domains, particularly medicine, are inherently multimodal. The integration of linguistic reasoning with visual perception has led to the development of Vision-Language Models (VLMs), which bridge the gap between structured textual knowledge and unstructured pixel data. By aligning these modalities in a shared latent space, VLMs facilitate a more holistic understanding of complex environments.

In the clinical domain, the utility of VLMs is transformative. Medical diagnostics rarely rely on a single data source; rather, they require the synthesis of medical imaging (e.g., MRI, CT, X-ray) and clinical narratives (e.g., radiology reports, electronic health records).  Specialized medical VLMs, such as Med-Flamingo (\cite{medflamingo}) and LLaVA-Med (\cite{medllava}), utilize contrastive learning or generative prefix-tuning to achieve semantic alignment between imaging features and medical terminology. This enables high-order tasks such as automated medical report generation, visual question answering (VQA) for clinical decision support, and cross-modal retrieval. By leveraging pre-trained knowledge from general-domain datasets and refining it with domain-specific corpora, these models address the critical challenge of "data silos" and expert-label scarcity in healthcare (\cite{medvlm}).

This review provides a systematic taxonomy of recent advancements in medical vision-language foundation models. We begin by dissecting the underlying architectures—categorizing them by their fusion strategies and backbone structures. Furthermore, we analyze key clinical applications, including diagnostic classification, semantic segmentation, and interactive VQA. Finally, we address the prevailing challenges including hallucinatory outputs, data privacy, and the ``domain gap" between natural and medical images.

\vspace{-4em}
\section{Vision-Language Multimodal Foundation Models}
VLMs represent a significant advancement in the integration of computer vision and natural language processing, enabling AI systems to perform a wide range of vision tasks with unprecedented accuracy in the medical domain. These models are designed to simultaneously comprehend and process visual and textual data, bridging the gap between visual and linguistic understanding. By combining image- and text-encoders, VLMs can extract meaningful features from medical images and associate them with relevant clinical information.

{\hyperref[fig:fig1]{\textbf{Figure 1}}} presents a generic block diagram of a foundation model, illustrating how different input data modalities are processed by multiple encoders, and their encoded representations are fused with text tokens before being passed into a large language model backbone. This integrated design governs the foundation models' outputs, enabling a wide range of applications such as visual question answering, image captioning, and medical diagnosis. In this section, we delve into notable research efforts aimed at integrating different types of foundation models into medical applications, showcasing their transformative potential in healthcare. We specifically discuss recent advancements categorized based on the architectures of foundation models, each bringing unique capabilities and challenges to the field. 

{\hyperref[fig:fig2]{\textbf{Figure 2}}} illustrates the temporal trends across different model categories, highlighting the growth in model size (in terms of parameters), the scale of training data, and the evolution of dataset availability over time. It also visualizes the distribution and size of datasets associated with each architectural category. In addition, {\hyperref[tab:models_combined]{\textbf{Table 1}} provide a comprehensive overview of the foundation models developed within each category, detailing their target backbones and supported tasks, including classification, segmentation, report generation, and question answering.



\begin{figure*}
    \centering
    \includegraphics[width=\textwidth]{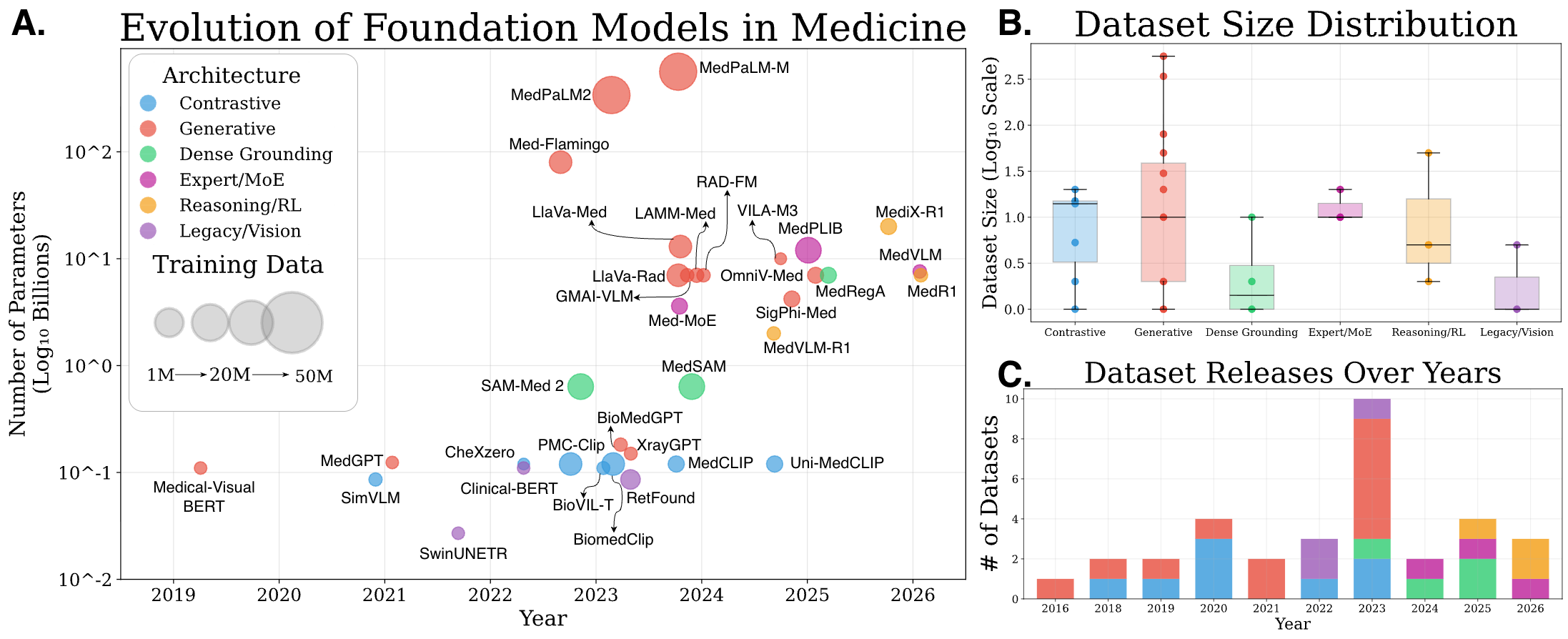}
    \caption{\textbf{A.} Evolution of foundation models over time, showing the relationship between the progress and model size. The size of each data point indicates the training dataset size, demonstrating the varying computational and data requirements across different models. \textbf{B.} Distribution of dataset sizes across different foundation model architectures, highlighting the substantial variation in them. \textbf{C.} Dataset releases over the years across architectures, revealing that considerable datasets have been released in recent years.}
    \vspace{-1em}
    \label{fig:fig2}
\end{figure*}

{\subsection{The Contrastive Alignment Paradigm}

The contrastive architectural paradigm, popularized by the Contrastive Language-Image Pre-training (CLIP) framework \cite{clip}, focuses on establishing a shared latent space where visual and textual representations are semantically aligned. The core mechanism involves a dual-encoder architecture that employs a contrastive loss function to maximize the cosine similarity between matched image-text pairs while minimizing the similarity for all other mismatched pairs in a given training batch. These models are highly prized for their zero-shot capabilities, allowing them to classify pathologies or retrieve relevant clinical cases based purely on natural language descriptions without requiring fine-tuning on labeled examples for every new condition. 

\subsubsection{Dual-Encoder Foundations and Adaptation Strategies}

BiomedCLIP stands as a preeminent example of the contrastive paradigm specifically tailored for the biomedical domain. Released by Microsoft Research in 2023, BiomedCLIP \cite{biomedclip} addresses the fundamental``domain gap" that renders web-trained VLMs ineffective in clinical settings. While the original CLIP utilizes a general-domain GPT-2 text encoder, BiomedCLIP is adapted from the CLIP framework by replacing the language backbone with PubMedBERT \cite{pubmedbert}. PubMedBERT is a language model pre-trained on the vast corpus of biomedical literature in PubMed, enabling the text encoder to natively understand clinical nomenclature, anatomical relations, and pathological nuances. For its vision backbone, BiomedCLIP utilizes a Vision Transformer \cite{vit} (specifically the ViT-L/16 architecture) adapted to handle higher image resolutions (up to 336 pixels) and captures fine-grained visual details essential for medical diagnosis. BiomedCLIP was not trained from scratch in the traditional sense but is an advanced adaptation of the CLIP approach, pre-trained on the PMC-15M dataset \cite{pmc15m}, a public collection of 15 million image-text pairs extracted from 4.4 million scientific articles. }

{MedCLIP \cite{medclip} further refines the contrastive paradigm by addressing the ``false negative" problem inherent in standard CLIP training. In clinical datasets, reports from different patients may describe identical findings (e.g., ``normal chest radiograph"), yet a standard contrastive loss would treat an image from Patient A and a report from Patient B as a negative pair even if they are semantically identical. MedCLIP utilizes a dual-encoder architecture with Swin Transformer \cite{swintransformer} or ResNet \cite{resnet} backbones for vision and domain-specific BERT for text, adapted from pre-trained checkpoints. The model’s defining innovation is the decoupling of images and texts for pre-training, which scales usable data in a combinatorial manner, and the implementation of a semantic matching loss that leverages medical ontologies like UMLS to soft-label pairs.}
{

\subsubsection{Specialized Radiology and Temporal Contrastive Variants}

CheXzero \cite{chexzero} represents a highly specialized application of contrastive learning for chest radiograph interpretation. It adopts the CLIP architecture using a ViT-B/16 vision backbone and a Transformer-based text encoder. CheXzero is an adaptation of general-domain CLIP weights, fine-tuned on approximately 200,000 pairs of chest X-ray images and radiology reports from the MIMIC-CXR \cite{mimic} dataset. This adaptation strategy allowed the model to match the diagnostic accuracy of human radiologists in a zero-shot multi-label classification setting, demonstrating that explicit labels are unnecessary when high-quality clinical reports are available for self-supervised alignment.}
{
BioViL-T \cite{biovitl} introduces temporal awareness into the contrastive paradigm, recognizing that radiologists' workflows often involve comparing current scans with historical baselines. The model utilizes a hybrid vision backbone composed of a ResNet-50 \cite{resnet} for initial feature extraction from images at each time point and a Vision Transformer to aggregate these features across the temporal dimension. Its text backbone is CXR-BERT-general \cite{cxrbert}, which is first pre-trained from scratch via masked language modeling on PubMed abstracts and clinical notes and then continually pre-trained for multimodal alignment. This model is essentially an adaptation of BERT and ResNet/ViT components, explicitly accounting for prior images and reports during both training and fine-tuning to achieve state-of-the-art performance on progression classification and phrase grounding.}
\newcolumntype{Y}{>{\raggedright\arraybackslash}X}
\newcolumntype{C}{>{\centering\arraybackslash}X}

\begin{table*}[!htbp]
\centering
\scalebox{0.7}{
\small
\begin{tabularx}{1.5\linewidth}{@{\extracolsep{\fill}} c Y c Y Y c @{}}
\toprule
\textbf{Architecture} & \textbf{Model (Year)} & \textbf{Params} & \textbf{Backbones (Vision / Language)} & \textbf{Training Status} & \textbf{Tasks} \\
\midrule

\textbf{Contrastive}
& SimVLM (2021) & $\sim$86M & ViT / Transformer LM & Adapted & C,R,Q \\
& CheXzero (2022) & $\sim$120M & ViT-B/16 / Transformer & Adapted (CLIP) & C \\
& BiomedCLIP (2023) & $\sim$120M & ViT-L/16 / PubMedBERT & Adapted (CLIP) & C,S,R,Q \\
& BioViL-T (2023) & $\sim$110M & ResNet-50 + ViT / CXR-BERT & Mixed (Scratch Text) & C \\
& PMC-CLIP (2023) & $\sim$120M & ViT-B/16 / PubMedBERT & Adapted (CLIP) & C,Q \\
& MedCLIP (2024) & $\sim$120M & Swin-T or ResNet / Domain BERT & Adapted & C,S,R,Q \\
& UniMed-CLIP (2025) & $\sim$120M & MetaCLIP ViT-B / BioMed-BERT & Adapted & C,S,R,Q \\

\midrule

\textbf{Generative}
& Medical-VisualBERT (2019) & $\sim$110M & CNN / VisualBERT & Adapted & C,Q \\
& MedGPT (2021) & $\sim$124M & Transformer LM & Scratch-trained & Q \\
& MedPaLM2 (2023) & $\sim$340B & Med-PaLM / PaLM & Adapted & C,R,Q \\
& Med-Flamingo (2023) & $\sim$80B & Flamingo Vision Encoder / LLM & Adapted & C,R,Q \\
& XrayGPT (2023) & $\sim$150M & MedCLIP / Vicuna & Adapted Encoders & C,Q \\
& BiomedGPT (2023) & $\sim$182M & BERT-style / GPT-style & Adapted & C,R,Q \\
& LLaVA-Med (2024) & $\sim$13B & CLIP ViT-L/14 / Vicuna-7B & Adapted (LLaVA) & Q \\
& Med-PaLM M (2024) & $\sim$562B & ViT / PaLM & Adapted (PaLM) & C,S,R,Q \\
& LLaVA-Rad (2024) & $\sim$7B & BiomedCLIP-CXR / Vicuna & Adapted & R,Q \\
& RadFM (2024) & $\sim$7B & Swin Transformer / LLaMA & Adapted & R,Q \\
& LAMM-Med (2024) & $\sim$7B & CLIP ViT-L / LLaMA-2 & Adapted & C,R,Q \\
& GMAI-VLM (2024) & $\sim$7B & ViT / LLaMA-based LLM & Adapted & C,R,Q \\
& SigPhi-Med (2025) & 4.2B & CLIP / Phi-2 & Adapted & C,S,R,Q \\
& OmniV-Med (2025) & $\sim$7B & RoPE-SigLIP / LLM backbone & Adapted & C,R,Q \\
& VILA-M3 (2025) & $\sim$8B--13B & ViT / LLaMA-3 & Adapted & C,R,Q \\

\midrule

\textbf{Dense Grounding}
& SAM-MED 2D (2023) & $\sim$636M & SAM ViT Encoder / N/A & Adapted (SAM) & S \\
& MedSAM (2024) & $\sim$636M & ViT / N/A & Adapted (SAM) & S \\
& VISTA3D (2025) & -- & SegResNet / Transformer & Mixed (Trained) & S \\
& MedRegA (2025) & $\sim$7B & InternVL variant / LLM & Adapted & C,S,R,Q \\

\midrule

\textbf{Expert/MoE}
& Med-MoE (2024) & $\sim$3.6B & CLIP ViT-L/14 / Phi-2 & Adapted (LLaVA) & C,S,R,Q \\
& MedPLIB (2025) & $\sim$12B & ViT-L/336 + ViT / MoE LLM & Adapted (CLIP/SAM) & C,S,R,Q \\
& Med3DVLM (2026) & $\sim$7.6B & DCFormer / ClinicalBERT & Adapted & C,S,R,Q \\

\midrule

\textbf{Reasoning/RL}
& MedVLM-R1 (2025) & $\sim$2B & Qwen2-VL-2B (Unified) & Adapted (RL) & Q \\
& Med-R1 (2026) & $\sim$7B & Qwen2-VL (Unified) & Adapted (RL) & Q \\
& MediX-R1 (2026) & $\sim$8B--30B & Baseline MLLM (Unified) & Adapted (RL) & C,R,Q \\

\midrule

\textbf{Legacy/Vision}
& SwinUNETR (2022) & $\sim$27M & Swin + UNET / N/A & Scratch-trained & S \\
& Clinical-BERT (2022) & $\sim$110M & BERT / N/A & Adapted & Q \\
& RETFound (2023) & $\sim$86M & ViT-Large / N/A & Scratch (SSL) & C \\

\bottomrule
\end{tabularx}}
\caption{Overview of medical foundation models grouped by architectural paradigm and ordered chronologically within each category. The table summarizes model scale, backbone architectures, training strategy, and supported downstream tasks including classification (C), segmentation (S), report generation (R), and visual question answering (Q).}
\label{tab:models_combined}
\end{table*}

{

\subsection{The Generative Multimodal Paradigm}

The generative architectural paradigm, or the Multimodal Large Language Model (MLLM) framework, integrates high-capacity vision encoders with powerful large language model (LLM) decoders to facilitate interactive clinical reasoning and open-ended conversation. Unlike contrastive models that produce simple similarity scores, generative Med-VLMs utilize an ``encoder-connector-decoder" structure where visual tokens are projected into the embedding space of the LLM, allowing the model to ``see" images as part of its prompt stream. This facilitates complex diagnostic assistance, structured report generation, and multi-turn visual dialogues that ground LLM logic in specific radiologic or pathologic findings. }
{

\subsubsection{LLaVA-based Architectures and Curriculum Learning}

LLaVA-Med \cite{medllava} is the seminal generative model for biomedicine, representing an adaptation of the general-domain Large Language and Vision Assistant (LLaVA). It utilizes a CLIP ViT-L/14 vision backbone to extract patch-level embeddings and a 7B-parameter LLM backbone, typically Vicuna-7B \cite{vicuna} or LLaMA \cite{llama}. LLaVA-Med employs a two-stage curriculum learning method: Stage 1 involves concept alignment by training a lightweight linear projection layer while keeping the vision encoder and LLM frozen; Stage 2 involves full-blown instruction-tuning of the LLM and the projector using GPT-4 generated multimodal dialogues. This adaptation process allows LLaVA-Med to transition from a general-purpose model to a specialized medical visual chatbot in less than 15 hours of training on a single 8-GPU node.
}
{
Further architectural diversification is seen in models like LLaVA-Rad \cite{llavarad} and CXR-LLaVA \cite{llavacxr}. LLaVA-Rad is adapted by integrating a domain-specific vision encoder, BiomedCLIP-CXR (pre-trained on 697,000 chest X-rays), with a Vicuna-7B-v1.5 language backbone using an MLP projector. In contrast, CXR-LLaVA's creators developed a custom image encoder (ViT-L/16) from scratch to handle grayscale chest radiograph textures specifically, which is then integrated with the LLaMA-2 7B language model through a linear multimodal projector that maps 1024-dimensional tokens into a 4096-dimensional space.
}
{
\subsubsection{Modular Generative Assistants and Multi-Agent Frameworks}

XrayGPT \cite{xraygpt} exemplifies a modular adaptation strategy by aligning a pre-trained specialized medical vision encoder (MedCLIP) with a fine-tuned medical LLM (Vicuna) using a simple linear transformation layer. The model focuses on chest radiograph summarization, leveraging approximately 217,000 interactive summaries generated from radiology reports to ground its conversational semantics.
}

{
Med-PaLM M \cite{medpalmm}, developed by Google, is a large-scale generalist multimodal system built upon the Pathways Language Model (PaLM) \cite{medpalm} architecture. It integrates visual information through two Vision Transformer (ViT) models of different parameter sizes—pre-trained on a classification dataset of 4 billion images—and utilizes the PaLM-E framework to handle interleaved sequences of multimodal inputs, including clinical language, imaging, and genomics, using the same set of model weights. Med-PaLM 2 \cite{medpalm2} further refines this by leveraging base model improvements (PaLM 2) and medical domain fine-tuning to achieve expert-level performance on USMLE-style questions and clinical topic datasets.
SigPhi-Med \cite{sigphimed} proposes a lightweight vision–language assistant for biomedical applications by replacing large language models with a smaller language backbone utilizing the Phi-2 backbone for vision. 
}
{

\subsection{The Dense Grounding \& Pixel-Level Paradigm}
\subsubsection{The Promptable Foundation Models}

As Med-VLMs advance, there is a shift from holistic image-level understanding toward fine-grained pixel-level perception and grounding. This architectural paradigm enables ``promptable" systems where models can respond to specific coordinates, bounding boxes, or textual references to identify anatomical regions and lesions. These models often utilize specialized mask decoders or region-aware encoders to bridge the gap between abstract diagnostic language and the spatial reality of clinical scans.
}

{
%
%
MedSAM \cite{medsam} is a fundamental adaptation of the Segment Anything Model (SAM) \cite{sam} for the medical domain. It retains the original ViT encoder and mask decoder architecture but is fine-tuned on a colossal dataset of over 1.5 million image-mask pairs covering 10 imaging modalities and 30 cancer types. This adaptation allows MedSAM to delineate anatomical structures and pathological findings across diverse modalities—from ultrasound to histopathology—with minimal user prompting. VISTA3D \cite{vista3d} is a unified foundation model designed specifically for 3D medical imaging segmentation, addressing the limitations of 2D slice-by-slice processing. Its vision backbone is based on SegResNet, integrated with a transformer-based prompt encoder that accepts class indices or 3D point clicks. VISTA3D utilizes a novel 3D supervoxel method to distill knowledge from 2D pre-trained backbones (like SAM) into a 3D volumetric context. The model is trained on 11,454 CT volumes using a multi-stage recipe that systematically trains automatic segmentation for 127 supported classes and zero-shot interactive segmentation for novel structures.
}
{

\subsubsection{Multi-Granular Pixel Grounding and Region-Aware Assistant}

MedPLIB \cite{medplib} introduces an end-to-end MLLM architecture capable of pixel-level visual grounding and reasoning. Its visual encoder is a dual-stream design: a CLIP ViT-L/336 vision tower for semantic features and a Visual Transformer with medical adapter layers (from SAM-Med2D) for pixel-level features. MedPLIB integrates these features with a Mixture-of-Experts (MoE) LLM that employs specialized experts for VQA and pixel grounding tasks. The model supports arbitrary pixel-level prompts—points, bounding boxes, and free-form shapes—and introduces a ``" token to identify features necessary for generating segmentation masks via a specialized mask decoder. MedRegA \cite{medrega} and Anatomy-VLM \cite{anatomyvlm} focus on region-aware alignment to mimic human-centric diagnostic workflows. MedRegA is a bilingual generalist model adapted from powerful general-domain backbones (e.g., InternVL), fine-tuned on the MedRegInstruct \cite{medrega} dataset to handle tasks like grounded report generation and text-to-region detection across 8 modalities. Anatomy-VLM employs an object detection encoder to localize anatomical structures and pathology features, which are then enriched with structured knowledge descriptions before being aligned via anatomy-wise contrastive learning.
}
\newcolumntype{Y}{>{\raggedright\arraybackslash}X} 
\newcolumntype{C}{>{\centering\arraybackslash}X}  

\begin{table*}[!htbp]
\centering   
\scalebox{1}{
\scriptsize  
\begin{tabularx}{\linewidth}{@{\extracolsep{\fill}} Y c Y C Y @{} } 
\toprule 
Dataset & Year & Size & Open-source & Tasks \\
\midrule

BraTS (\cite{brats}) & 2020 & $\sim10K$ images & $\checkmark$ & S \\

CheXpert (\cite{chexpert}) & 2019 & $\sim224K$ chest X-rays & $\checkmark$ & C \\

MS-CXR-T (\cite{mscxrt}) & 2022 & 1162 images & $\checkmark$ & C \\

$21K$ medical images (\cite{sam2dmed}) & 2023 & $21K$ images & $\checkmark$ & S \\

CX-XHR (\cite{cxxhr}) & 2020 & 85,184 images & $\checkmark$ & C \\

COVID-19 CT (\cite{covidct}) & 2020 & 40,175 images & $\checkmark$ & C \\

MedMD (\cite{medmd}) & 2023 & $16M$ samples & $\times$ & C,S,R \\

MEH-MIDAS (\cite{mehmidas}) & 2023 & 37,401 images & $\checkmark$ & C \\

RSNA Pneumonia (\cite{rsna}) & 2018 & $\sim30K$ chest X-rays & $\checkmark$ & C,S \\

Colossal Clean Crawled Corpus (\cite{simvlm}) & 2023 & -- & $\checkmark$ & C,S,R \\

MIMIC-III (\cite{mimiciii}) & 2016 & $\sim58K$ ICU admissions & $\checkmark$ & R,Q \\

MultiMedQA (\cite{medpalmm}) & 2023 & 12,723 QA pairs & $\checkmark$ & Q \\

MTB (\cite{medflamingo}) & 2023 & -- & $\checkmark$ & R,Q \\

PMC-OA (\cite{pmcoa}) & 2023 & $\sim3.9M$ full-text articles & $\checkmark$ & R,Q \\

MIMIC-CXR (\cite{mimic}) & 2019 & $\sim377K$ images, $\sim227K$ reports & $\checkmark$ & R \\

VQA-RAD (\cite{vqarad}) & 2018 & $\sim315$ images, $3K$ QA pairs & $\checkmark$ & Q \\

ALIGN (\cite{simvlm}) & 2021 & $\sim1.8B$ image-text pairs & $\checkmark$ & C,S,R \\

PMC-15M (\cite{pmc15m}) & 2023 & $15M$ figure-caption pairs & $\checkmark$ & C,S,R,Q \\

{Path-VQA (\cite{pathvqa})} & {2020} & {$\sim5K$ images, $\sim33K$ QA pairs} & {$\checkmark$} & {C,Q} \\

{SLAKE (\cite{slake})} & {2021} & {$\sim642$ images, $\sim14K$ QA pairs} & {$\checkmark$} & {C,S,Q} \\

\bottomrule
\end{tabularx}}
\caption{Summary of datasets used in foundation models for medicine. Tasks are abbreviated as C (Classification), S (Segmentation), R (Report Generation), and Q (Question Answering).}
\label{tab:datasets_unified}
\end{table*}
{

\subsection{The Expert-Centric \& Sparse Computation Paradigm}
\subsubsection{The Modular Mixture-of-Experts Paradigm}

To mitigate the computational costs of scaling Med-VLMs while enhancing their ability to handle heterogeneous clinical data, researchers have adopted Mixture-of-Experts (MoE) and modular architectural paradigms. These models replace traditional dense feed-forward networks with a set of sparsely activated experts, where a routing mechanism dynamically selects the most appropriate expert for a given input—such as an MRI expert for neuroimaging or a pathology expert for histology slides. 
}
{
Med-MoE \cite{medmoe} is a lightweight framework designed to reduce parameter activation by 30\%–50\% while maintaining performance comparable to much larger models. Utilizing small-scale LLM backbones like Phi-2 (2.7B) or StableLM (1.7B), Med-MoE incorporates multiple domain-specific experts along with an ``always-on" meta-expert that captures global medical information. The model is an adaptation of LLaVA-style frameworks, where a modality-aware router is pre-trained to identify imaging types (CT, MRI, pathology, X-ray) and direct the token stream to the relevant expert. MedMoE focuses specifically on adapting local visual representations based on the diagnostic context and report type. It routes multi-scale features through expert branches conditioned on the accompanying clinical narrative. Using a Swin Transformer backbone to derive feature pyramids, MedMoE enables spatially adaptive attention that emphasizes regions of the image aligned with the semantic granularity of the associated report.
}
{

\subsubsection{Modular 3D Volumetric Processing}

Med3DVLM \cite{med3dvlm} addresses the inherent complexity of 3D volumetric data through architectural modularity. It integrates three key innovations: (1) DCFormer, an efficient encoder that uses decomposed 3D convolutions to capture volumetric features with reduced complexity; (2) a SigLIP-based sigmoid language-image pre-training scheme for stable alignment; and (3) an MLP-Mixer projector that effectively fuses low- and high-level image features with LLM text embeddings. Med3DVLM is an adaptation of the LLaVA framework, utilizing the DCFormer-small variant and ClinicalBERT \cite{clinicalbert} as backbones, pre-trained on 120,084 3D medical images.The table below details the routing and expert configurations of prominent MoE and modular models.}
{

\subsection{The Generative Diffusion \& Reconstruction Paradigm}

The diffusion paradigm replaces autoregressive decoding with iterative denoising, offering a flexible alternative for text and image generation. DiffuMamba (or DiffuApriel) is a masked diffusion language model built on a bidirectional Mamba-2 backbone, matching Transformer-based diffusion quality with up to 8.2x higher throughput on long sequences. In image generation, RoentGen \cite{roentgen} is an adaptation of the Stable Diffusion (SD) pipeline, fine-tuned on real chest X-rays to synthesize high-fidelity scans conditioned on textual reports. MSTG (Mamba-Spatial-Temporal Generator) integrates CNNs and Mamba within a diffusion probabilistic model to generate 4D temporal medical images, effectively modeling spatial and sequential dependencies.}
{
\subsection{The Reasoning-Incentivized / RL Paradigm}

A major criticism of perception-oriented Med-VLMs is their lack of sophisticated reasoning capabilities required for real-world clinical contexts. Supervised fine-tuning often results in models that memorize surface patterns rather than developing robust understanding. To address this ``reasoning gap," researchers have introduced reinforcement learning (RL) frameworks designed to incentivize step-by-step logic and clinical guideline alignment. MedVLM-R1 \cite{medvlmr1} and Med-R1 \cite{medr1} utilize Group Relative Policy Optimization (GRPO) to cultivate emergent reasoning capabilities without requiring expensive, explicit reasoning annotations. MedVLM-R1 is adapted from the Qwen2-VL-2B \cite{qwen2vl} backbone and is fine-tuned on radiological images (MRI, CT, X-ray) to generate explicit  rationales alongside final answer selections. This model demonstrates that compact architectures can outperform generalist giants like Qwen2-VL-72B by focusing on rule-based rewards for structural compliance and accuracy. Med-R1 also leverages the Qwen2-VL backbone and GRPO, supporting 8 diverse imaging modalities and key diagnostic tasks like lesion grading and biological attribute analysis. Med-R1 explores the nuances of the "thinking" process, revealing that for certain cross-domain tasks, omitting intermediate rationales can actually improve generalization by reducing overfitting to specific reasoning styles. }
{

\subsection{The Data Infrastructure}

The technical success of these architectural paradigms is inextricably linked to the curation of large-scale biomedical datasets. PMC-15M \cite{pmc15m} has emerged as the global standard for contrastive and generative pre-training, providing 15 million image-text pairs extracted from publicly available scientific articles. This dataset solves the ``supervision bottleneck" by moving beyond the narrow focus of radiology to include pathology, microscopy, and biomedical illustrations, enabling the development of generalist Med-VLMs like BiomedGPT and Med-PaLM M. MIMIC-CXR \cite{mimic} remains the definitive resource for radiology-specific models like CheXzero, BioViL-T, and CXR-LLaVA. Its combination of high-resolution DICOM images and detailed free-text reports allows models to learn the fine-grained relationship between visual evidence and clinical impression. The recent release of M3D-Data further expands this infrastructure to 3D volumetric analysis, providing 120,084 CT/MRI volumes paired with radiology reports for models like Med3DVLM. {\hyperref[tab:datasets_unified]{\textbf{Table 2}}} summarizes the commonly used datasets for medical VLMs along with their task based utility. 
}

\section{Application of Medical VLMs}

\subsection{Visual Question Answering \& Report Generation}

One common application of foundation models and medical vision-language models is their integration into chatbots, facilitating interactive question answering and clinical report generation. These foundation models have demonstrated promising performance and assistive capabilities in generating clinical reports. For instance, ClinicalBERT (\cite{clinicalbert}) and RadioBERT (\cite{radiobert}), which build upon the BERT architecture, exhibit remarkable effectiveness in disease prediction and report generation based on user-input symptoms and test results. 

While BERT is widely used for medical chatbots and report generation, GPT-based models have gained traction following the success of ChatGPT. Models such as GPT-2, GPT-3.5, and GPT-4 (\cite{chatgpt}) have been employed to generate radiology reports based solely on patient information and imaging findings, with their performance validated by practicing radiologists, indicating their applicability in clinical settings. 

Additionally, BiomedGPT (\cite{biomedgpt}) is a GPT-based model extensively fine-tuned for medical tasks, enabling end-to-end question answering and report generation. BiomedGPT attains the highest accuracy for question answering and has a decent performance for report generation.

Another significant effort involves analyzing the performance of existing open-source chatbots like ChatGPT and Claude for question-answering. Studies indicate that ChatGPT and Claude can extract valuable information from clinical notes and patient data, providing relevant and accurate answers. These advancements underscore the potential of foundation models in improving medical chatbot functionality and clinical workflows.

\begin{table*}[!htbp]
\centering
\scriptsize
\scalebox{1}{
\begin{tabularx}{\textwidth}{@{\extracolsep{\fill}} c l l l l @{}}
\toprule
\textbf{Year} & \textbf{Model} & \textbf{Modality} & \textbf{Task} & \textbf{Organ} \\
\midrule

\multirow{3}{*}{2022} 
& RoentGen (\cite{roentgen}) & X-ray & Generation & Chest \\
& ChexZero (\cite{chexzero}) & X-ray & Interpretation & Chest \\
\midrule

\multirow{3}{*}{2023} 
& RadFM (\cite{radfm}) & Radiology & Interpretation & Tissue \\
& RetFound (\cite{Retfound}) & Ophthalmology & Interpretation & Eye \\
& X-RayGPT (\cite{xraygpt}) & X-ray & Interpretation & Chest \\
& PLIP (\cite{plip}) & Histopathology & Analysis & Tissue \\
\midrule

\multirow{7}{*}{2024} 
& CONCH (\cite{pathfm}) & Histopathology & Classification & Tissue \\
& EchoCLIP (\cite{echo}) & Echocardiogram & Interpretation & Heart \\
& Merlin (\cite{merlin}) & CT & 3D Tomography & Abdomen \\
& CheXagent (\cite{chexagent}) & X-ray & Interpretation & Chest \\
& VisionUnite (\cite{visionunite}) & Ophthalmology & Multimodal analysis & Eye \\
& EyeCLIP (\cite{eyeclip}) & Ophthalmology & Interpretation & Eye \\
\midrule

\multirow{3}{*}{2025} 
& FLAIR (\cite{flair}) & Ophthalmology & Interpretation & Eye \\
& MUSK (\cite{musk}) & Whole-body Imaging & Classification & General \\
& NasVLM (\cite{nasal}) & Nasal Endoscopy & Diagnosis & Nose \\

\bottomrule
\end{tabularx}
}
\caption{Domain-specific vision-language foundation models categorized by year, modality, task, and organ system.}
\label{tab:domain_models}
\end{table*}

\subsection{Pathology Classification}

One of the most common applications of deep learning models in computer vision is classification. 
Generalized classification models typically rely on heavily annotated datasets, which are often scarce in the medical domain. Consequently, VLMs targeting classification tasks employ zero-shot or few-shot learning as viable alternatives. 
For instance, CheXzero (\cite{chexzero}) has demonstrated strong performance in pathology classification on chest X-ray (CXR) (\cite{chexpert}) images using single-shot learning with minimal annotations, achieving accuracy comparable to that of radiologists. CheXzero attains a high AUC score across multiple diagnostic tasks.
Similarly, CXR-CLIP (\cite{mscxrt}) utilizes image-text pairs along with general prompts for both classification and report generation.
Other strategies to address the need for extensively labeled images in medical classification include optimized fine-tuning methods that leverage positive pair loss to enhance performance alongside zero-shot learning. The widespread adoption of zero- or few-shot learning techniques for medical image classification has resulted in powerful VLMs being effectively utilized for not only question answering and report generation but also for classification tasks. This evolution underscores the potential of these models to overcome challenges associated with limited annotated datasets in the medical field, thereby improving diagnostic accuracy and efficiency.

\subsection{Medical Image Segmentation}

Medical image segmentation is a crucial aspect of medical diagnostics, playing a key role in non-invasive information retrieval, visualization, and analysis. Several models, including MedSAM (\cite{medsam}) and Swin UNETR (\cite{swinunetr}), have fine-tuned existing state-of-the-art architectures like SAM and U-Net for medical imaging applications. These models have demonstrated remarkable performance not only in larger object segmentation but also in achieving benchmark results for polyp segmentation. The ViT-based Swin UNETR model, for example, has achieved the highest scores in the BraTS 2021 (\cite{brats}) segmentation challenge, highlighting its effectiveness in brain tumor segmentation. Both MedSAM and Swin UNETR achieve comparable F1 scores on medical segmentation tasks, with SAM-Med2D performing closely behind.

Another significant model is nnU-Net (\cite{nnunet}),  which automates the configuration of U-Net for various medical imaging tasks, optimizing performance across diverse datasets. Furthermore, the use of Generative Adversarial Networks (GANs) in models like CycleGAN (\cite{cyclegan}) has shown promise in tasks such as domain adaptation and enhancing image quality for segmentation purposes.

While significant progress has been made in developing VLMs and corresponding datasets for 2D medical image segmentation, recent research is increasingly focusing on extending these capabilities to 3D segmentation tasks (\cite{sam3d}). Vista3D (\cite{vista3d}) presents a unified foundation model for both 2D and 3D medical image segmentation, emphasizing versatility across modalities while minimizing the need for extensive 3D annotations. To address the scarcity of large-scale annotated 3D datasets, Mosaic3D (\cite{mosaic3d}) was recently introduced. It combines a 3D encoder trained via contrastive learning with a lightweight mask decoder, enabling open-vocabulary semantic and instance segmentation in 3D medical volumes. In parallel, SAM-Med3D (\cite{sammed3d}) proposes a cross-dimensional adaptation strategy that leverages 2D SAM features for effective 3D segmentation, highlighting the potential of 2D-to-3D knowledge transfer. These advancements underscore the growing emphasis on generalizable 3D foundation models in medical imaging, particularly in data-scarce environments.

Overall, these advancements in medical image segmentation highlight the potential of both traditional and deep learning-based methods to improve diagnostic accuracy and facilitate better clinical decision-making across various imaging modalities.

\section{Bridging the Gap Between Foundation Models and Real-World Healthcare Practitioners}

The integration of foundation models into healthcare presents a transformative opportunity to improve clinical decision-making and streamline workflows. While these models have demonstrated substantial promise in tasks such as diagnostics, report generation, and medical image analysis, their successful deployment in real-world clinical settings requires not only technical proficiency but also alignment with clinical workflows and healthcare-specific requirements (\cite{foundation}).

Vision-language multimodal foundation models are designed to generalize across diverse medical imaging modalities and tasks by jointly learning from visual and textual inputs. However, their broad scope often limits clinical applicability due to the inherent heterogeneity in imaging techniques, anatomical structures, and diagnostic requirements. Consequently, there has been a growing focus on developing specialized vision-language models that are optimized for particular clinical domains, leveraging targeted datasets and task-specific objectives. {\hyperref[tab:domain_models]{\textbf{Table 3}}} summarizes key vision-language foundation models designed for domain-specific medical applications.

CONCH (\cite{pathfm}), for example, is a vision-language foundation model developed for histopathology. Trained on a diverse dataset that includes histopathology images, biomedical text, and over 1.17 million image–captions pairs, CONCH achieves state-of-the-art performance on classification, segmentation, and captioning tasks across 14 histology datasets. Similarly, PLIP (\cite{plip}) is a vision-language model trained on pathology images from OpenPath, showing a 2.5\% improvement in F1 score over prior baselines and excelling in image-text retrieval and classification tasks.

EchoCLIP (\cite{echo}) is designed for echocardiography, leveraging ultrasound images and expert interpretations to facilitate automated and accurate echocardiogram analysis. Merlin (\cite{merlin}), a 3D vision-language model, integrates structured electronic health records (EHRs) with unstructured radiology reports for the interpretation of abdominal CT scans. The model has shown strong performance in internal validation and real-world data settings.

RoentGen (\cite{roentgen}) is a vision-language model for chest X-ray (CXR) generation. It synthesizes realistic, diverse CXR images controllable via free-form radiological text prompts, expanding the potential for data augmentation and medical image simulation. Similarly focused on chest X-ray analysis, CheXagent (\cite{chexagent}) specializes in disease classification and diagnosis, emphasizing model transparency and decision interpretability.

In ophthalmology, FLAIR (\cite{flair}) and EyeCLIP (\cite{eyeclip}) address retinal and general eye disease diagnosis, respectively, utilizing multimodal training data to enable image-text alignment for tasks such as disease classification and report generation. VisionUnite also focuses on ophthalmology, incorporating multimodal representations to support diagnosis across a broader range of eye conditions.

In the context of oncology, MUSK (\cite{musk}) is a domain-specific foundation model trained for cancer diagnosis and treatment prediction, leveraging both imaging and clinical text. NasVLM (\cite{nasal}) addresses nasal disease diagnosis by integrating ENT-specific medical images and clinical descriptions, facilitating automated classification and report generation in rhinology.

These specialized models build upon general-purpose foundation architectures and adapt them to meet the unique demands of specific clinical domains. While they demonstrate superior performance on curated benchmarks and in-domain datasets, their translation to real-world deployment is still hindered by challenges such as generalization across institutions, lack of prospective validation, and integration into clinical decision support systems.

In real-world settings, medical practitioners do not rely on a single source of information; instead, they utilize multiple images, reports, and test results to arrive at accurate diagnoses. Collaborations between AI developers and clinicians are essential to create user-friendly interfaces that seamlessly integrate FMs into existing systems, ensuring that these tools support rather than disrupt clinical decision-making (\cite{future}). {With their ability to process and reason across multimodal data including imaging, text-based reports, and structured clinical records FMs are increasingly being integrated into real-world healthcare environments. However, effectively leveraging these systems requires substantial training and curriculum development to ensure that practitioners understand both their capabilities and limitations \cite{llm_assit, AI_Assit}. When appropriately integrated and supported with adequate training, studies suggest that FM-assisted workflows can significantly enhance clinical performance, with reports indicating improvements of up to 27.5\% in diagnostic reasoning scores \cite{rtc_1}. This highlights the potential of FMs not only as automated tools but as collaborative decision-support systems that augment clinicians’ analytical capabilities.}

{Beyond assisting physicians in diagnosis and clinical decision-making, medical foundation models are increasingly being applied to human activity recognition (HAR) tasks to support remote patient monitoring \cite{har}. By analyzing multimodal data streams—such as wearable sensor signals, video, and physiological measurements—these models can detect patient behaviors, track rehabilitation progress, and identify abnormal activities such as falls or mobility impairments. Such capabilities make remote monitoring systems more informative and proactive, enabling clinicians to intervene earlier and provide more personalized care, particularly for elderly patients or individuals with chronic conditions. }

{Although foundation models (FMs) have demonstrated significant capabilities in both research and clinical settings, they are not without limitations. Notably, issues such as hallucinations and embedded biases remain important concerns. Hallucinations may lead models to generate plausible but incorrect medical information, while biases arising from imbalanced or non-representative training data can affect the reliability and fairness of model predictions. Although substantial research efforts are underway to mitigate these challenges through improved training strategies, better dataset curation, and enhanced evaluation frameworks \cite{hallucination}, these issues have not yet been fully resolved. Consequently, the integration of FMs into real-world clinical applications must be approached with caution, ensuring appropriate safeguards, validation protocols, and human oversight. }
By addressing these challenges and fostering collaboration between clinicians, researchers, and technology developers, we can bridge the gap between foundation models and their practical application in healthcare, ultimately improving patient outcomes and enhancing the efficiency of medical services.
\begin{figure}
  \vspace{-1em}
    \centering
    \includegraphics[scale = 0.65]{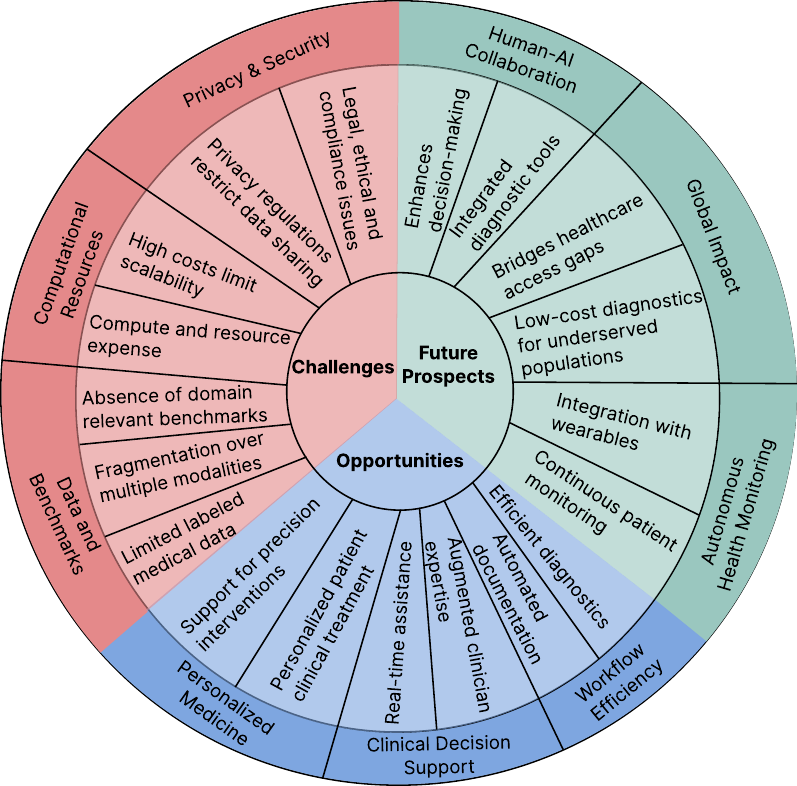}
    \caption{Overview of the Challenges, Opportunities, and Future Prospects for Foundation Model in the medical domain.}
    \vspace{-1em}
    \label{fig:fig4}
\end{figure}
\section{Challenges and Opportunities}



{The transition of vision-language foundation models into clinical practice is fundamentally constrained by the high-dimensional complexity of medical data and the stringent reliability requirements of safety-critical environments. A primary algorithmic challenge lies in the scarcity of diverse, high-quality annotated datasets for rare pathologies, which often results in models that struggle with the "long-tail" distribution of diseases. While generative frameworks like RoentGen attempt to address this through the synthesis of realistic, text-controllable radiographs for data augmentation, ensuring the anatomical fidelity of synthesized pixels across heterogeneous imaging modalities remains a significant technical hurdle. Furthermore, the inherent black-box nature of large-scale transformer architectures limits their interpretability, as standard attention maps frequently fail to provide a medically grounded justification for a model's diagnostic output. }

{
Recent technical shifts toward region-specific grounding, such as those implemented in MedRegA, represent a necessary move toward identifying fine-grained anatomical structures to improve model transparency and clinical trust.The deployment of these models also faces substantial systemic and computational obstacles. Privacy regulations such as HIPAA and GDPR restrict the large-scale data aggregation required to train dense models, necessitating the exploration of decentralized paradigms like Federated Learning and the implementation of Differential Privacy layers to facilitate cross-institutional collaboration. }

{
Computationally, the quadratic complexity of self-attention mechanisms poses a significant barrier to processing high-resolution 3D volumetric data, such as CT or MRI scans, in real-time clinical workflows. To mitigate these resource requirements, the field is increasingly adopting sparse architectures like the Mixture-of-Experts (MoE) framework utilized in Med-MoE and MedPLIB, which selectively activates only a subset of parameter experts during inference. These efficient designs, coupled with the emergence of lightweight "Small Language Models" like SigPhi-Med \cite{sigphimed}, offer a viable path for democratizing expert-level AI diagnostics in resource-constrained environments where massive compute infrastructure is unavailable.}

{
Looking toward the horizon of General-purpose Medical AI (GMAI)\cite{fm_survey, survey}, the integration of reinforcement learning represents a critical evolution in enhancing the logical reasoning of multimodal systems. Emerging paradigms such as Group Relative Policy Optimization (GRPO), employed by models like Med-R1, incentivize logical reasoning paths over simple linguistic pattern matching, thereby reducing the risk of clinically dangerous hallucinations. This shift enables a future where foundation models act as agentic clinical assistants capable of autonomous health monitoring and precision medicine. By synthesizing multi-sensory inputs—including imaging, structured electronic health records, and genomic data—into a unified multimodal latent space, these models could facilitate highly personalized treatment ecosystems tailored to individual genetic profiles. The successful realization of this future vision hinges on the technical refinement of these models to handle complex clinical inquiries with the consistency and precision required to augment, rather than replace, human expertise.}

{\hyperref[fig:fig4]{\textbf{Figure 3}}} summarizes the challenges, opportunities, and future prospects of the foundation model in the medical domain. 



\section{Conclusion}

In this review article, we explored the evolution of foundation models in the medical domain, highlighting their growing impact on various healthcare applications. We examined the development of general and vision-language models, the datasets that fuel their training, and their role in key downstream tasks such as medical image segmentation, disease classification, report generation, and clinical question answering. While these models demonstrate remarkable potential in automating and enhancing medical workflows, challenges remain in terms of data heterogeneity, model interpretability, domain adaptation, and regulatory considerations. Addressing these challenges will require continued advancements in model design, data curation, and validation frameworks. Moving forward, interdisciplinary collaboration between AI researchers, clinicians, and regulatory bodies will be essential to ensure the safe, effective, and ethical deployment of foundation models in real-world medical settings, ultimately improving patient outcomes and the efficiency of healthcare delivery.

This research did not receive any specific grant from funding agencies in the public, commercial, or not-for-profit sectors.

\bibliographystyle{unsrtnat}
\bibliography{references}

%



\end{document}